\documentclass[conference]{IEEEtran}
\ifCLASSINFOpdf
\else
\fi
\usepackage{url}

\usepackage{xcolor}
\usepackage[pdftex]{graphicx}
\usepackage{amsfonts}
\usepackage{amsmath}
\usepackage{paralist}
\usepackage{cite}
\usepackage{wrapfig}
\usepackage[utf8]{inputenc}
\usepackage[english]{babel}
\usepackage{amsthm}
\usepackage{subfig}
\usepackage{graphicx}
\usepackage{enumitem}
\usepackage{mdframed}

\newtheorem{definition}{Definition}

\newcommand{\aka}{a.k.a.~}

\begin{document}
%
\title{Prescriptive Process Monitoring for \\ Cost-Aware Cycle Time Reduction}


\author{
    \IEEEauthorblockN{Zahra Dasht Bozorgi\IEEEauthorrefmark{1}, 
    Irene Teinemaa\IEEEauthorrefmark{2},
    Marlon Dumas\IEEEauthorrefmark{3}, 
    Marcello La Rosa\IEEEauthorrefmark{1}, 
    Artem Polyvyanyy\IEEEauthorrefmark{1}}
    \IEEEauthorblockA{\IEEEauthorrefmark{1}The University of Melbourne, Australia
    \\zdashtbozorg@student.unimelb.edu.au, \{marcello.larosa,artem.polyvyanyy\}@unimelb.edu.au}
        \IEEEauthorblockA{\IEEEauthorrefmark{2}Booking.com, The Netherlands
    \\irene.teinemaa@booking.com}
    \IEEEauthorblockA{\IEEEauthorrefmark{3}University of Tartu, Estonia
    \\marlon.dumas@ut.ee}
}


%


\maketitle

\begin{abstract}
Reducing cycle time is a recurrent concern in the field of business process management. Depending on the process, various interventions may be triggered to reduce the cycle time of a case, for example, using a faster shipping service in an order-to-delivery process or calling a customer to obtain missing information rather than waiting passively. However, each of these interventions comes with a cost. This paper tackles the problem of determining if and when to trigger a time-reducing intervention in a way that maximizes a net gain function. The paper proposes a prescriptive monitoring method that uses orthogonal random forests to estimate the causal effect of triggering a time-reducing intervention for each ongoing case of a process. Based on this estimate, the method triggers interventions according to a user-defined policy. The method is evaluated on two real-life datasets. 
\end{abstract}



%
\IEEEpeerreviewmaketitle

\section{Introduction}
\noindent Reducing cycle time (i.e.\ the time spanned between the start and completion of a case) is a recurrent improvement objective in the field of business process management~\cite{DBLP:books/sp/DumasRMR18}. Depending on the process, there may be several \emph{interventions} (a.k.a.\  \emph{treatments}) that workers may perform to reduce cycle time. For example, in an application-to-approval process, giving a call to a customer to obtain missing information may speed up a case. However, such interventions come at a cost, and they can only be done if a suitable worker is available.

In this setting, this paper tackles the following problem: \textit{Given an intervention that in general reduces the cycle time of a case, for which cases and at which point in the execution of a case, should we trigger this intervention to maximise the total net gain?} Here, the total net gain is the sum of the differences between the benefit of each intervention and its cost.
To tackle this problem, we propose a prescriptive monitoring method that triggers interventions based on continuous observation of the process. The method relies on a causal effect estimation technique, namely orthogonal random forests, to estimate the effect of an intervention on the cycle time of a case. When the estimated causal effect exceeds a user-defined threshold (the \emph{policy}), an intervention is triggered. The method incorporates an approach to select the policy based on estimated net gain. We evaluate the method using two real-life datasets.



The next section motivates the approach via an example. Sect.~\ref{sec: related work} reviews related work. Sect.~\ref{sec: preliminaries} introduces preliminary concepts and notations. The proposed method is described in Sect.~\ref{sec:proposed approach}, while the evaluation is discussed in Sect.~\ref{sec: evaluation}. Sect.~\ref{sec: conclusion} draws conclusions and discusses future work directions.

\section{Motivating Example}
\label{sec: example}
\noindent We consider a loan origination process that starts when a client submits a loan application. The submitted documents are screened. If some documents are missing, a request for further information is sent to the customer either by email or phone. 
Handling missing document events via email takes less effort. Historical data, though, shows that customers respond faster to phone calls than to emails, thus reducing the cycle time. But given that phone calls are more costly, workers would rather only call a customer if this would speed up the process. 

One approach to determine which customers to call (and when) is by predicting the remaining time of each case and triggering an intervention on those cases that are expected to be most delayed w.r.t.\ a cycle time target~\cite{teinemaa2018alarm,MetzgerNBP19,metzger2020triggering}. However, this approach may be ineffective. Suppose that a case is likely to have a long cycle time because the employee handling it is busy. If so, making a phone call to obtain the missing information  (the intervention) would consume more resources with little effect on cycle time. 
A more suitable approach is to estimate the effect of a phone call on the cycle time of each case and to direct the interventions to cases with the highest  estimated effect. This paper pursues this approach.

\section{Related Work}
\label{sec: related work}
\subsection{Prescriptive Process Monitoring}
\noindent Various prescriptive process monitoring methods have been proposed. Teinemaa~\emph{et~al}. \cite{teinemaa2018alarm} propose to trigger an intervention when the probability of a case leading to a negative outcome is above a threshold optimised w.r.t.\ a net gain function. The method by Weinzierl~\emph{et~al}.~\cite{weinzierl2020prescriptive} uses predictive models to determine which activity, among the most likely next activities in the case, is correlated with higher values of a given KPI. Metzger~\emph{et~al}.~\cite{metzger2020triggering} use deep learning models to generate predictions about an ongoing case and feed these predictions to an online reinforcement learning technique, which triggers an intervention based on the predictions and their reliability. The above methods tackle the problem of identifying which cases need an intervention, while our approach aims to identify cases that can benefit most from a given intervention.

\subsection{Causal Inference in Process Mining}
\noindent Causal inference has been widely applied in the field of process mining.
Hompes~\emph{et~al}.~\cite{DBLP:conf/caise/HompesMRDBA17} propose an approach to discover cause-effect relations between aggregate characteristics of a process (e.g.\ frequency of activity) and process performance indicators (e.g.\ mean cycle time). Koorn~\emph{et~al}.~\cite{DBLP:conf/bpm/KoornLLR20} present a method to identify causal relations between a response and an effect. They use statistical tests to discover action-response-effects, where the action defines a subgroup of cases and the response is a treatment that enhances the probability of the effect.
Polyvyanyy~\emph{et~al}.~\cite{POLYVYANYY2019345} present causality mining, a systematic approach to discovering causal dependencies between events encoded in large datasets.  Narendra~\emph{et~al}.~\cite{DBLP:conf/bpm/NarendraA0D19} use structural causal models to confirm potential cause-effect relations identified by analysts. 
Qafari~\emph{et~al}.~\cite{DBLP:conf/bpm/QafariA20} use structural equation models to test for the existence of a causal relation between an attribute and an outcome. They later expand the use of SEMs in \cite{qafari2021case} for counterfactual explanation of case-level predictions. 
The above studies do not quantify the effect of a treatment at the level of an individual case. Hence, they cannot be applied to individualized prescriptive process monitoring, which is the focus of this paper.


In previous work~\cite{DBLP:conf/icpm/BozorgiTDRP20}, we use action rule mining to extract candidate treatments correlated with a positive outcome. 
That work deals only with binary case outcomes, whereas in this paper we deal with a numerical target variable (cycle time). 

\section{Preliminaries and Definitions}
\label{sec: preliminaries}

\subsection{Event logs and traces}
\label{subsec: math&pm}


\noindent 
We refer to an instance of a process execution as a \textit{case}. 
A case consists of a set of \textit{events}, where an event represents the execution of an activity. 
Each event has three attributes: 
a \textit{case identifier} specifying which case the event belongs to,
an \textit{activity} that triggered the event, and 
a \textit{timestamp} specifying when the event occurred. 
An event may have further attributes, such as a resource that carried out the activity or an event type. 

\begin{definition}[Event, Trace, Event Log]
\label{def:trace event log}
\emph{An \emph{event} is a tuple $(a,c,t,(d_1, v_1),\ldots,(d_m, v_m))$, $m \in \mathbb{N}_0$, where 
$a$ is an activity name (label), 
$c$ is a case identifier, 
$t$ is a timestamp, and 
$(d_1, v_1) \ldots, (d_m, v_m)$ are attribute-value pairs.
A \emph{trace} is a finite sequence $\sigma = \langle e_1,\ldots,e_{n} \rangle$, $n \in \mathbb{N}_0$, of events with the same case identifier. 
An \emph{event log} $L$ is a multiset of traces.}
\end{definition}


As we aim to estimate the causal effect of an intervention on the cycle time of a case, we are interested in events that occurred before the intervention. We use the notion of a k-prefix to capture such preceding events.

\begin{definition}[k-Prefix]
\label{def: prefix}
\emph{A \emph{k-prefix} of a trace $\langle e_1,\ldots,e_{n} \rangle$, $n \in \mathbb{N}_0$
is a sequence $\langle e_1,\dots, e_{k} \rangle$, $0 \leq k \leq n$.
}
\end{definition}


\subsection{Causal Inference and Causal Machine Learning}
\label{subsec: ci&ml}
\noindent
The field of causal inference is concerned with determining the independent effect of a phenomenon on an outcome of interest. 
Two causal inference frameworks are discussed in the literature. 
In the \emph{causal graphical models} framework, a causal graph is constructed by a domain expert. Given a causal graph, it is possible to ascertain whether a causal estimand is identifiable and if so, it can be estimated using automated methods.
While this framework is focused on the identification of causal effects, the \emph{Neyman-Rubin potential outcomes} framework is focused on the estimation of causal effects of interventions.
Hence, in this paper, we use the \emph{Neyman-Rubin potential outcomes} framework.

An intervention (\aka\ \emph{treatment}) is represented by a binary variable $T \in \{0,1\}$, where $T=1$ indicates that the treatment is applied and $T=0$ that it is not. 
Each case in a log has two potential outcomes: the outcome under treatment ($Y^{1}$) and the outcome under no treatment ($Y^{0}$). Given this, the \emph{conditional average treatment effect} (CATE) is defined as follows.

\begin{definition}[Conditional Average Treatment Effect]
\emph{Let $X$ be a set of attributes that characterise a case. 
\vspace{-1mm}
\begin{center}$\mathit{CATE}: \theta(x) = \mathbb{E}[Y^{1}-Y^{0} \,|\, X=x]$\end{center}
}
\end{definition}
\vspace{-1mm}

CATE is a causal estimand, meaning that in order to estimate it, we need to have access to both $Y^{0}$ and $Y^{1}$. 
However, in real-life, it is impossible to follow both potential realities. 
This is known as the ``fundamental problem of causal inference''~\cite{holland1986statistics}.
Thus, to estimate the causal estimand, we express it via statistical estimands, such as $\mathbb{E}[Y\,|\,T\!=\!1,X]$ and $\mathbb{E}[Y\,|\,T\!=\!0,X]$, which can be estimated from data. 
According to the potential outcomes framework, CATE can be expressed via statistical estimands only if the \emph{exchangeability}, \emph{positivity}, \emph{consistency}, and \emph{no interference} conditions hold.

The exchangeability (\aka ignorablility) condition means that given the pre-treatment attributes $X$, treatment assignment is independent of the potential outcomes. 
In other words, after conditioning on $X$, the treatment assignment should be as good as random, which ensures that the treated and not treated groups are exchangeable, that is:
\vspace{-1mm}
\begin{center}
$Y^1,Y^0 \perp \!\!\! \perp T\,|\,X$.
\end{center}
\vspace{-1mm}
The positivity condition means that for every set of values for $X$, treatment assignment is not deterministic. So, every subgroup of interest has some chance of getting either treatment:
\begin{center}
$P(T=t \,|\, X=x)>0$, for all $t$ and for all $x$.
\end{center}
\vspace{-1mm}
The consistency condition states that the potential outcome under treatment $T=t$ is equal to the observed outcome if the actual treatment received is $T=t$:
\vspace{-1mm}
\begin{center}
$Y=Y^t$ if $T=t$ for all $t$.
\end{center}
\vspace{-1mm}
According to no interference, the potential outcomes of one subject are not affected by the treatments received by others.

The positivity and consistency conditions can be verified from data and by ensuring that there are no multiple definitions of the treatment under study. However, verification of the exchangeability condition is not straightforward.
In observational data, such as event logs, there often exist variables that influence both the treatment assignment and the outcome.
The existence of these \emph{confounding variables} (or \emph{controls}) creates a non-causal association between $T$ and $Y$, which can invalidate the study. 
The best way to circumvent this problem is to conduct a randomised experiment (A/B test).
As we are working with event logs, where randomisation of the treatment is not ensured, we \emph{assume} that exchangeability holds. 
Under this assumption, the observed variables $X$ contain sufficient information needed to adjust for confounding. The adjustment can then be carried out during the estimation step.


In many real-life use cases, including business processes, the no interference condition can often be violated as well. For instance, skipping an activity in one case might leave the process worker available to perform activities in another case, and vice versa. As a result, the potential outcomes of the second instance are affected by the treatment applied in the first. However, it is common to assume that a violation of the no interference condition has only a small influence on the causal effect estimates and is considered negligible in practice.

Next to identification, the other core problem in causal inference is estimating CATE from observational data. A large body of works from recent years focus on using machine learning methods for CATE estimation. 
A popular method is the work by Athey~\emph{et~al}.~\cite{athey2019generalized}. which is a flexible non-parametric estimation method based on \emph{generalised random forests}.
However, it does not allow for high dimensional set of confounders. 
This problem is addressed in a method called \emph{double machine learning} proposed by Chernozhukov~\emph{et~al}.~\cite{10.1111/ectj.12097}. Oprescu~\emph{et~al}.~\cite{DBLP:conf/icml/OprescuSW19} generalise the ideas from these two approaches in an approach called \emph{orthogonal random forests}.

\subsection{Orthogonal Random Forests}

\noindent We use orthogonal random forests (ORFs) to estimate the effects of treatments on cases. When applied to the problem of heterogeneous treatment effect estimation, ORFs assume the data $D$ to be in the form $D = \{(T_i, Y_i, W_i, X_i)\}^{n}_{i=1}$, for $n$ observations. For each observation $i$, $T_i$ is the received treatment, $Y_i$ is the observed outcome, $W_i$ represents potential confounding variables, and $X_i$ are the features capturing heterogeneity. The ORF method makes the assumptions captured in the following structural equations:
\vspace{-1mm}
\begin{center}$Y = \theta(X)\cdot T + f(X,W) + \epsilon$,\end{center}
\begin{center}$T = g(X,W) + \eta$,\end{center}
\vspace{-1mm}

\noindent where $f(X, W)$ models the outcome $Y$ when no treatment is applied, $\theta(X)$ is the treatment effect function (CATE), and $g(X,W)$ captures the relationship between treatment $T$, confounders $W$, features $X$, and $\epsilon$ and $\eta$ are unobserved noises, such that $E[\epsilon \,|\, X, W, T] = 0$ and $E[\eta \,|\, X, W, \epsilon] = 0$. 

ORF follows the residualisation approach used in double machine learning. Double machine learning, as the name suggests, consists of two stages. In the first stage, a model is fit to predict $Y$ from $X$ and $W$, and another model is fit to predict $T$ from $X$ and $W$, that is:
\vspace{-1mm}
\begin{center}
$\hat{Y} = \mathbb{E}[Y \,|\, X,W]$ \,\,\,and\,\,\, $\hat{T} = \mathbb{E}[T \,|\, X,W]$.
\end{center}
\vspace{-1mm}

In the second stage, the effects of confounding are removed by fitting a model to predict $Y-\hat{Y}$ from $T-\hat{T}$. The treatment effect function $\theta(x)$ is defined as follows:
\vspace{-1mm}
\begin{center}
$\mathbb{E}[Y - \hat{Y} - \theta(x)\cdot(T - \hat{T}) \,|\, X=x]=0$.
\end{center}
\vspace{-1mm}


\section{Approach}
\label{sec:proposed approach}

\begin{figure*}[t]
\vspace{-5mm}
	\centering
	\includegraphics[width=1\linewidth,height=3cm]{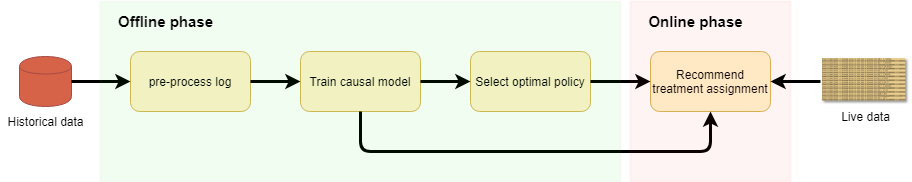}
	\caption{Overview of the proposed approach}
	\label{fig:framework}
	\vspace{-5mm}
\end{figure*}

\noindent We aim to design a recommender system that seeks to minimise the cycle time of a process case in a cost-aware manner. At the core of our system is a causal estimation module that estimates the effect of a given treatment on a target metric. 
The target is to decrease the cycle time of a process case. 
We assume that the treatments are binary. 
For instance, suppose that in the loan origination process discussed in Sect.~\ref{sec: example}, the default way of handling missing documents is by sending emails to clients to request additional documents, while some customers get phone calls instead.
In this scenario, the binary treatment options are sending emails ($T=0$) or making phone calls ($T=1$). We refer to cases that got a phone call as the treatment group and the cases that got an email as the control group. 
Making a phone call should, in general, speed up the process. 
However, it is not feasible to call every client because of the associated additional costs. 
Thus, we want to find the best policy for deciding which clients should get phone calls.


Our approach consists of two phases, as depicted in Figure~\ref{fig:framework}. In the \emph{offline phase}, case prefixes are extracted from the log and used to train a causal effect estimation model. 
Then, the best model threshold is selected. This threshold defines a treatment policy. In the \emph{online phase}, the causal effect estimator and the selected policy are used to  recommend whether to treat or not to treat an ongoing case based on its estimated treatment effect and the selected treatment policy.

Next, we describe the three steps of the offline phase, namely \emph{data pre-processing}, \emph{causal model construction}, and \emph{policy selection}), followed by the online phase.



\subsection{Data Pre-processing}
\label{subsec: data preprocessing}


\subsubsection{Data Cleaning}
First, we pre-process the log to remove incomplete cases. 
We then repair missing attribute values. 
For a numeric attribute, we set missing values to the median value of that attribute. 
For a categorical attribute, we set missing values to the most observed value for that attribute (the mode). 

\subsubsection{k-Prefix Extraction}
To construct feature vectors, we select a single prefix of length $k$ from each case. In cases where the treatment assignment is present in the data, $k$ is the length of the prefix before the actual treatment time. For instance, given a trace  $\sigma = \langle e_1,\ldots,e_{n} \rangle$, suppose that $e_{k+1}$ indicates the treatment event where $k+1 \leq n$. In this case, the selected prefix is $\langle e_1,\ldots,e_{k} \rangle$. Thus, $k$ can be different for different cases. In cases where the treatment did not occur, we estimate the treatment point by randomly drawing from the distribution of prefix lengths when the treatment is present. 

\subsubsection{Prefix Encoding}
To apply the machine learning method for estimating treatment effects, we need to encode trace prefixes (ongoing cases) as fixed-sized feature vectors. Encoding static case attributes is straightforward since their values do not change during the execution of a case. However, traces often contain dynamic event attributes whose values change as the case unfolds. Various methods have been proposed to encode dynamic attributes~\cite{teinemaa2019outcome}, including \emph{aggregation encoding}, \emph{last-state encoding}, \emph{index-based encoding}, and \emph{tensor encoding}~\cite{10.1145/3331449}. 
We use aggregation encoding for the activity type and resource attributes and last-state encoding for event attributes. 
Hence, the feature vector constructed from a k-prefix contains one numeric feature for each activity type or resource A, the number of times A appears in the k-prefix. For other event attributes (e.g.\ ``offered loan amount'' or ``type of loan offer''), the feature vector encodes only the last value of this attribute (e.g.\ the last offered loan amount). 
If an encoded attributes is a categorical attribute, we apply one-hot encoding to represent it as a numeric feature vector -- a common practice when using tree-based ensemble models, such as ORF.

\subsubsection{Temporal and Workload Features}
Given that we seek to estimate treatment effects on cycle time, we include temporal information in the feature vectors.
Specifically, we include the month, weekday, and hour of the timestamp of the last event in the prefix, the time between the first and the last event in the prefix, and the time between the last two events in the prefix (i.e.\ the inactivity period prior to the most recent event).

We also include the number of active cases as a feature to act as a proxy for the current workload in the process. We do so because  workload is a potential confounder, influencing both the cycle time and the treatments (during high workloads, cases may be delayed, and the effect of treatments might be weaker than usual). Finally, we encode the difference between the first event in the prefix (the start time of the case) and the first event in the log (the start time of the log's timeframe) since the process might behave differently at different points in time; hence, this feature can be a confounder.


\subsection{Causal Model Construction}
\label{subsec: CausalRecommendations}
This step of the offline phase takes as input the encoded k-prefixes and learns a function that, given a k-prefix $\langle \mathbf{x}^{(1)},\mathbf{x}^{(2)}, \dots, \mathbf{x}^{(k)} \rangle$, returns a point estimate of the treatment effect $\theta$ and the corresponding  confidence interval. 

We use ORFs to implement the causal model.
The use of ORFs has several advantages. ORFs support non-parametric estimation of the target variable while allowing for a high-dimensional set of confounding variables $W$.
This is useful in the context of process mining since traces typically have a large number of event attributes, which can lead to feature explosion. The problem is exacerbated when there is a large number of resources and categorical attributes. Finally, due to ORFs being asymptotically normal (i.e.\ the distribution of the estimated treatment effects approaches a normal distribution as the sample size grows), they allow for the construction of valid confidence intervals based on bootstrapping.

An ORF requires four inputs: the target variable $Y$, the treatment indicator $T$, features capturing heterogeneity $X$, and the confounding variables $W$. Since this recommender system aims to decrease cycle time, the outcome $Y$ is the cycle time of the process. We assume that a binary treatment is previously identified, which is hypothesised to cause a decrease in cycle time. In the loan origination example, this is a phone call.

As the proposed approach can be applied at an operational level, we would like to know the treatment effect when it is time for the process worker to decide whether to treat a case or not. Therefore, effect heterogeneity is captured via static case-level attributes and all event attributes (including the activity name and the resource), and the inter-case features available at the decision time. This means that $X$ is a feature vector that captures all of this information about the k-prefix after the suitable encodings have been applied, as described above. In this study, we assume that all of the available features derived from the event log are also potential confounders, meaning that all the features present in $X$ are included in $W$. However, in general, $X$ and $W$ do not need to be the same. A domain expert can remove some features from $W$ if she believes they do not influence the treatment decision or outcome.

Armed with these definitions of $Y$, $T$, $X$, and $W$, the next step is to train an ORF. The input data is temporally split into the train and test sets. A separate test set is required for the evaluation of the trained model, and for selecting the best policy. As mentioned in the preliminaries, we need to learn the treatment effect function $\theta$ by solving the same set of equations as in double machine learning. This is a two-phase approach that requires $Y$ and $T$ to be modeled locally at $X=x$, obtaining the estimates $\hat{Y}$ and $\hat{T}$. In the second stage, the treatment effect is estimated by minimising the residual of $Y$ (denoted by $Y - \hat{Y}$) on the residual of $T$ (denoted by $T - \hat{T}$) locally at $X=x$ using the squared loss:
\vspace{-1mm}
\begin{center} $\theta(x) = argmin_{\theta}\mathbb{E}[(Y-\hat{Y}-\theta\cdot(T-\hat{T}))^2 \,|\, X=x]$ \end{center}
\vspace{-1mm}

 ORFs allow the use of flexible models for the estimation of $\hat{T}$ and $\hat{Y}$. 
 These models can be chosen based on their performance on specific datasets. 
 Also, at prediction time, we have the option to use different models to residualise the treatment and the outcome. 
 Fitting $\hat{Y}$ and $\hat{T}$ locally around the target feature $X$ ensures that more weight is put on samples that are similar in the feature space.
 A random forest with a causal criterion is constructed to capture the similarity metric in the $X$ space, so that samples that are similar in the feature space $X$ have similar treatment effects. 

\subsection{Policy Selection}
\noindent 
Oftentimes, applying treatments to cases comes at a cost. Making phone calls to customers is an example of this. Having a treatment effect estimate for every running case can help decision makers separate the cases that would benefit from the treatment from those that would not be affected or would be negatively affected by it. Particularly, if the cost of the treatment is not far below from its benefit, it is important to carefully select a policy to determine which cases to treat. Accordingly, in this step, we propose a policy selection module that maximises the net benefit given the cost of the treatment and the treatment effect estimates. Furthermore, our proposed policy selection approach provides a way to evaluate the ORF model that is used to estimate the treatment effects. 

The first step in the policy selection module is to build a \emph{Qini curve}. Qini curves and the closely related \emph{uplift curves} provide a way to evaluate CATE estimators when the ground truth treatment effect is not available, which is the case in all real-world datasets~\cite{DBLP:journals/corr/abs-2007-12769}. Since our target variable in this study is continuous and the goal is to reduce it, we use a modified version of the Qini curve presented below.

Given an estimator $\hat{\theta}$ and k-prefixes $\sigma_i$ (with the treatment decision happening at the k-th event), let $\pi$ be the ascending ordering of the traces according to their estimated treatment effect, i.e., $\hat{\theta}^\pi(\sigma_i)\leq\hat{\theta}^\pi(\sigma_j), \forall\, i<j$. In addition, $\pi(n)$ is used to denote the first $n$ percent of traces from the ordering.

Let $R_{\pi(n)}$ be the sum of the duration of cases in $\pi(n)$, i.e., $R_{\pi(n)}=\sum_{i\in\pi(n)}Y_i$. Furthermore, let $R_{\pi(n)}^{T=1}$ and $R_{\pi(n)}^{T=0}$ be, respectively, the sums of the durations of cases in the treated and control groups in $\pi(n)$. Also, to denote the number of traces in the treated and control groups in $\pi(n)$, we use $N_{\pi(n)}^{T=1}$ and $N_{\pi(n)}^{T=0}$, respectively. We now define the Qini curve as:
\vspace{-2mm}
\begin{center}$Qini(n)=R_{\pi(n)}^{T=0}\times \frac{N_{\pi(n)}^{T=1}}{N_{\pi(n)}^{T=0}}-R_{\pi(n)}^{T=1}$.
\end{center}
\vspace{-1mm}

$Qini(n)$ is the expected total reduction in cycle time, given that the top $n$ percent of cases selected by the ORF model are treated (e.g. we make phone calls to the top $n$ percent of cases). The Qini curve can be plotted by computing $Qini(n)$ for different values of $n$. In addition to evaluating our ORF model, we use the Qini curve for policy selection by incorporating the cost and the benefit multipliers to create \emph{net value curves}. Suppose $v$ is the value of reducing the cycle time by one unit of time and $c$ is the cost of applying the treatment to one case. Then, the net gain of applying the treatment to $\pi(n)$ is defined as follows:
\vspace{-1mm}
\begin{center}
$gain(n)=v \times Qini(n) - c \times N_{\pi(n)}^{T=1}$.
\end{center}
\vspace{-1mm}

Similar to the Qini curve, the net value curve can be drawn by computing $gain(n)$ for different values of $n$. The user has the option to view the curve and to select the optimal policy based on organisational constraints. For example, suppose the organisation has a target of achieving a net benefit of $x$. In that case, the net value curve will provide the minimum proportion of the population that needs to be treated to achieve that goal. If there are no constraints, the policy that yields the highest net value can be selected automatically.

\subsection{Online phase}
\noindent In the online phase, the applicability of the treatment for an ongoing case with an observed k-prefix is assessed. If the treatment is not applicable at that point, the assessment is repeated after the next event. If the treatment is applicable, the k-prefix of the case is encoded as a feature vector using the same approach as in the offline phase. Then, the treatment effect is estimated using the pre-trained ORF model. If the estimated net gain is sufficiently high according to the pre-selected treatment policy, the prescriptive monitoring system recommends applying the treatment.

Let us come back to the example loan origination process. 
First, we take a log of this process, clean it, encode the case-level and event-level features, and create a treatment attribute, where the value of this attribute is zero for cases that got an email and 1 for cases that got a phone call. 
We then divide this pre-processed dataset into training and testing sets. We use the training set to train an ORF model. This trained model can then take an incomplete case as input and returns, as a number, the estimated effect of calling the customer. 
We then estimate the effect of the phone call on all the cases in the test set. We use these estimates to create a net-value curve. The policy maker uses the curve to decide the percentage of future cases that will get the phone call based on organisational constraints. 

In the online phase, a new application is created by the customer. After the execution of each activity, the applicability of the phone call at that stage is assessed. In the initial stages of document processing, there is no need for the phone call, so we move on to the next activity. If we encounter missing documents in this case, we encode the information about this case using the same method as in the offline phase and estimate the treatment effect using our trained ORF. Suppose that the selected policy is to treat half of the cases. If the estimated effect for this new case is in the top 50\% of the cases we have had so far, we call this customer; otherwise, we send an email.

\section{Evaluation}
\label{sec: evaluation}
\noindent To assess our approach, we conducted experiments on two real-life logs, measured the estimated causal effect of our recommendations, and compared the results with improvement suggestions identified by non-causal methods commonly used for prescriptive process monitoring~\cite{teinemaa2018alarm,MetzgerNBP19,metzger2020triggering}. 

\begin{table*}[t]
\vspace{10pt}
    \centering
    \vspace{-8mm}
\footnotesize{\begin{tabular}{|r|l|l|l|l|l|l|l|l|l|l|}
  \hline
  \textbf{Log}  & \textbf{Number of trees} & \textbf{Minimum leaf size} & \textbf{Maximum depth} & \textbf{Sub-sample ratio} & \textbf{Lambda reg} \\ 
  \hline
  
  BPI17 & 200     & 20 & 30  & 0.4 & 0.01 \\
  BPI19 & 200     & 10 & 10  & 0.7 & 0.01 \\
  \hline
\end{tabular} }
\caption{Parameter setting for training the ORF models}
\vspace{-5mm}
\label{table:parameters}
\end{table*}

\subsection{Experimental Setup}
\noindent The logs we used are the BPI17 and BPI19 logs, available from the 4TU Centre for Research Data.\footnote{\url{https://data.4tu.nl}} We used these two datasets because they contain interventions that could possibly have a causal effect on the cycle time. Moreover, each log shows unique features. BPI17 is characterised by a combination of case-level and event-level attributes, while most of the attributes in the BPI19 log are case-level. 


\vspace{3pt}
\noindent \textbf {BPI17}: This log contains traces of a loan application process of a Dutch financial institute. The data contains attributes about the applications and the loan offers made by the bank.  




\vspace{3pt}
\noindent \textbf{BPI19}: This log contains traces from a purchase-to-pay process of a Dutch multinational company. Each case describes a purchase order item from its creation to payment. This log also records activities such as changes, cancellations, and message exchange related to purchase orders. 

\noindent \textbf{Evaluation measures:} In the literature, CATE estimators are often evaluated by computing Qini curves. The underpinning intuition is that if the CATE is estimated accurately, the cases with a positive outcome in the treated group would have a higher estimated CATE than those in the same group with negative outcomes. Also, cases in the control group with a negative outcome should have a higher CATE than the positive outcome cases in the same group. Thus, a desirable causal model has a Qini curve above the random curve.

\noindent \textbf{Training setting:} We split the data into 60\%--20\%--20\% for training, validation, and testing, respectively, by preserving the temporal order between cases. We used the training set to train the ORF model, the validation set to tune the hyperparameters, and the test set to provide an unbiased evaluation of the model. The details of the training settings are shown in Table \ref{table:parameters}.

\subsection{Results}
\noindent We discuss the results obtained by running the ORF algorithm on the two logs above, and compare our results with non-causal methods based on two predictive models: Lasso and Random Forest (RF). We chose these two methods because they had high performance in predicting cycle time in these two logs. As shown in~\cite{DBLP:journals/corr/abs-2102-07298}, their performance on these logs is only slightly lower than deep adversarial models, which are the state-of-the-art in predictive monitoring, while having significantly lower training time. To resemble the setting assumed by predictive monitoring techniques, these models were trained on untreated cases only. Specifically, the baseline approach is one that predicts the cycle time given that no intervention is made, and triggers an intervention if the predicted cycle time is above a threshold. This threshold is set in such a way that X\% of the cases in the testing set are targeted, where X is an independent variable in the experiment.

\subsubsection{BPI17} This log shows a number of cases with unusually long duration. This severely impacts customer experience and, in many cases, leads to a negative outcome. However, since this is an interactive process, the cause of delay in many cases is that the bank is waiting for input from the customer to move on with the process. So, one possible intervention is to contact the customer to ask for additional information. While this intervention is effective for many of the cases observed in this log, the cost of calling all customers might be too high, or the company might not have the required resources to call everyone. So, we applied our approach to identify which cases benefit more from receiving this additional phone call. 

\subsubsection{BPI19} In this log, we observe some activities that indicate a change in the purchase order items. Since changes might lead to re-work, we hypothesised that avoiding changes leads to lower cycle times. Specifically, we considered skipping of activity \emph{Change Price} as our treatment for this experiment. However, fixing the price of purchase orders all at the same time for each case without considering the specific context of each case leads to a rigid order placement procedure and might require extra work at the beginning of the process. Thus, rather than fixing the price at the beginning of every case, it is beneficial to have a targeted approach. Furthermore, our approach provides a recommendation for each case, indicating whether a price change should be permitted or not. In this way, we can avoid the rework in cases where this change is harmful and increases cycle time while preserving flexibility in other cases that are not highly affected by price change. 

\subsubsection{Discussion} 
\begin{figure}[!h]
	\vspace{-\baselineskip}
	\centering
	\subfloat[BPI 2017]{\label{fig:qini17}\includegraphics[width=0.4\textwidth,height=4cm]{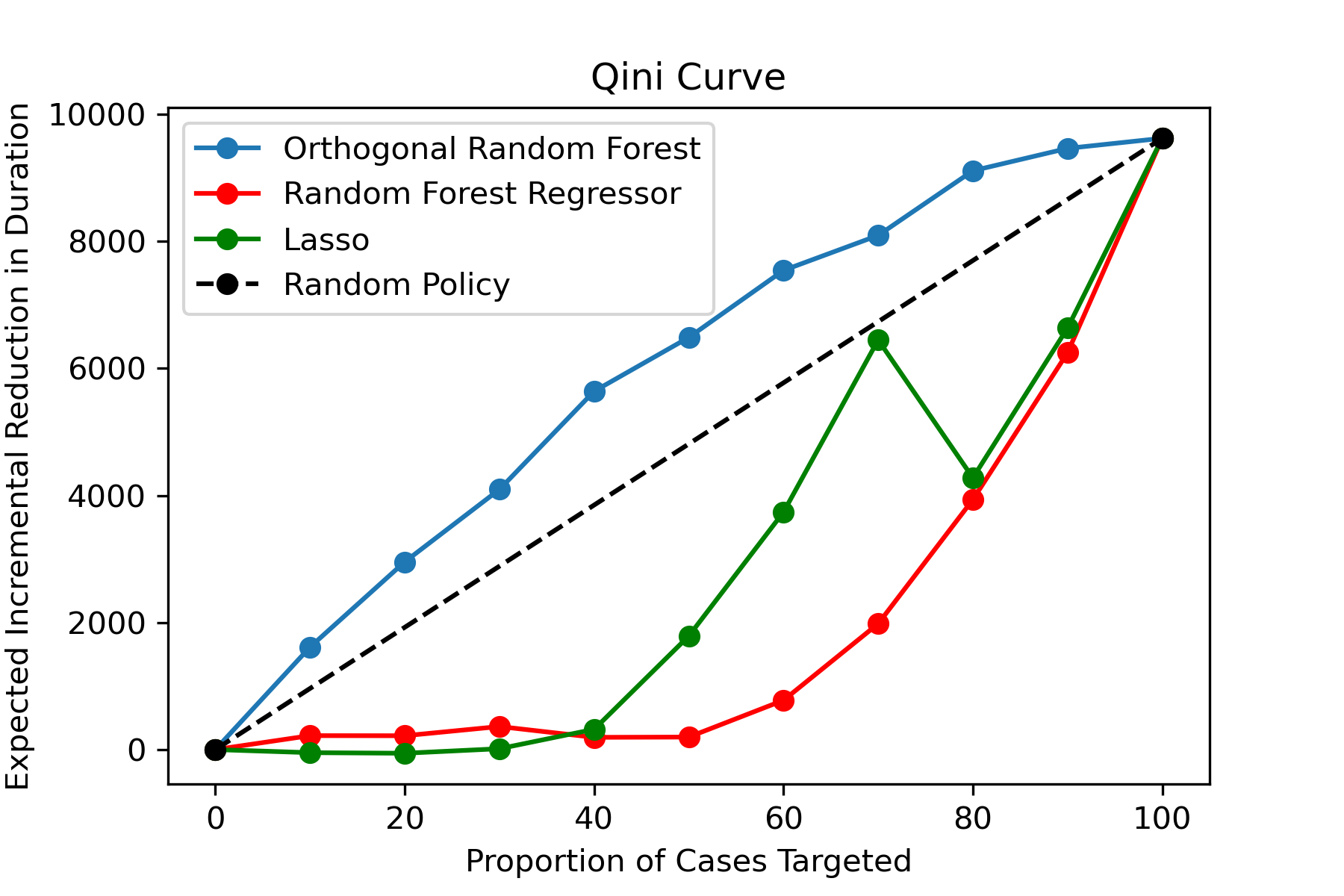}}\\
	\subfloat[BPI 2019]{\label{fig:qini19}\includegraphics[width=0.4\textwidth,height=4cm]{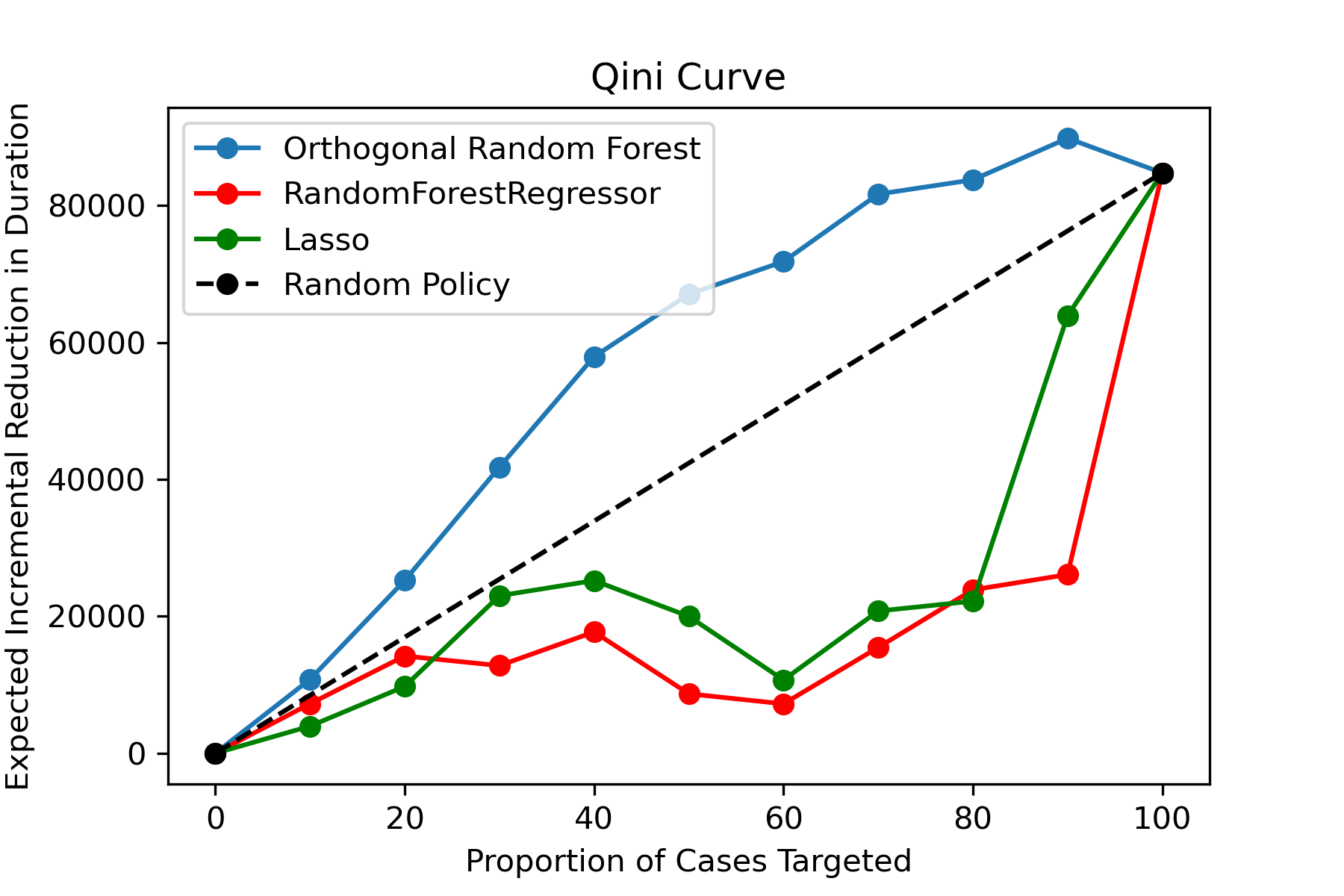}}
	\caption{Qini curves for ORF models.}
	\label{fig:QiniCurve}
\end{figure}

The plots in Figure \ref{fig:QiniCurve} show the Qini curves of the models constructed in our experiments. The black dashed line shows the expected incremental reduction in cycle time if a random policy was used to treat a certain percentage of the cases and the blue, red, and green lines show the reduction if the policy was based on the ORF, RF, and Lasso models respectively. It can be seen that in both datasets, the expected reduction is higher if the ORF-based policy is followed rather than policies based on non-causal predictive models. Notably, the curves for RF and Lasso are below the random policy line indicating that \textit{while predictive models are good at identifying which cases will take a long time, they are not necessarily good at identifying which cases should be targeted with the chosen treatments}. The Qini curve for the BPI17 log shows that treating 90\% of the cases with the highest treatment effect gives the same reduction as treating everybody. Particularly in the curve for BPI19, it can be seen that just treating 90\% of cases gives a higher reduction in cycle time than treating everybody. This curve is not monotonically increasing as usual cumulative gain charts because the quantity on the y-axis is the \emph{difference} between treated and non-treated cases, which in some segments can be negative since the treatment hurts the outcome in these cases. In the plot for BPI17, we do not see a decrease because the treatment does not hurt the outcome, but the treatment effect is low enough not to result in any gain in the cases in the last decile.

\begin{figure}[!h]
	\vspace{-\baselineskip}
	\centering
	\subfloat[BPI 2017]{\label{fig:v/c=1}\includegraphics[width=0.4\textwidth, height=4cm]{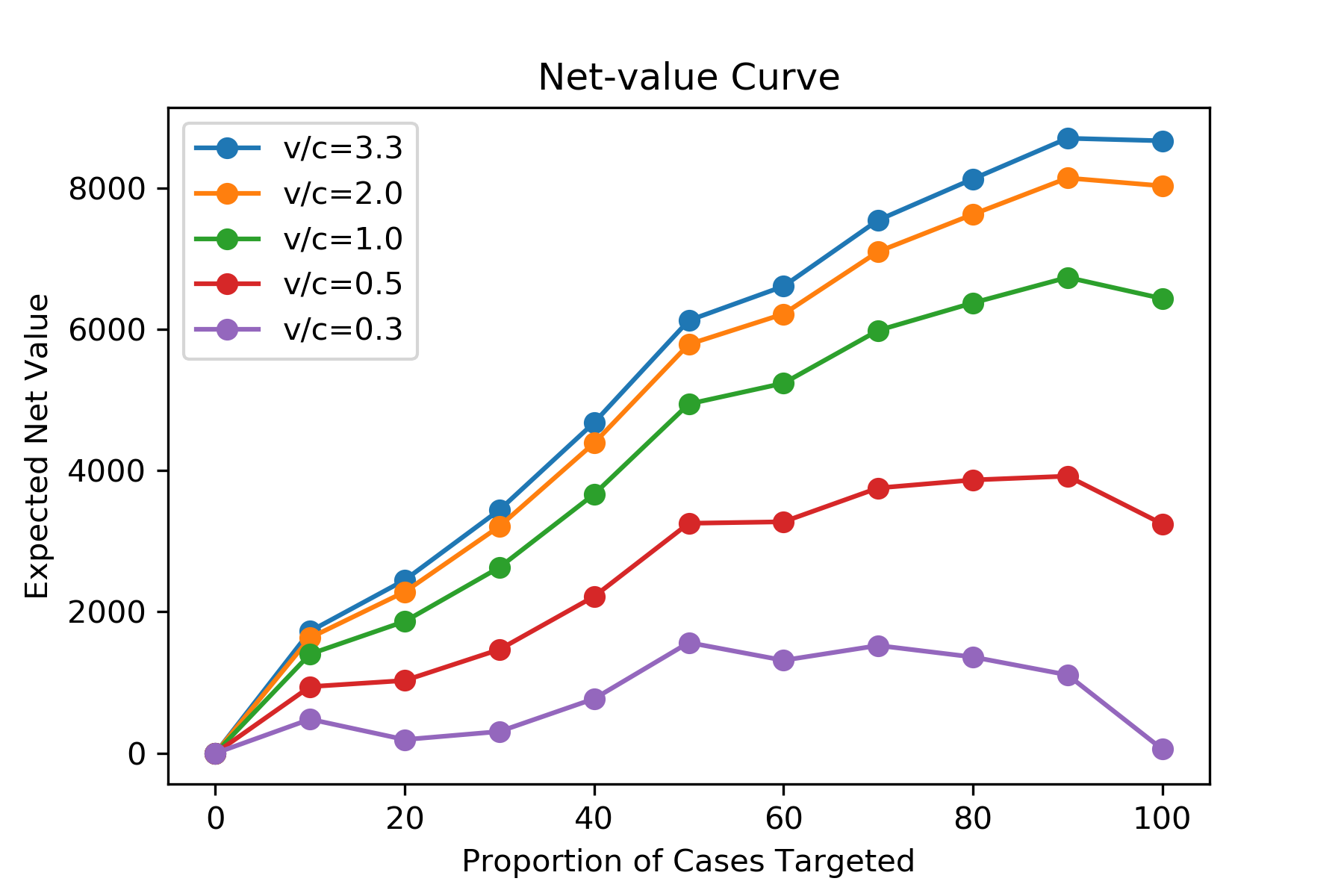}}\\
	\subfloat[BPI 2019]{\label{fig:v/c=2}\includegraphics[width=0.4\textwidth, height=4cm]{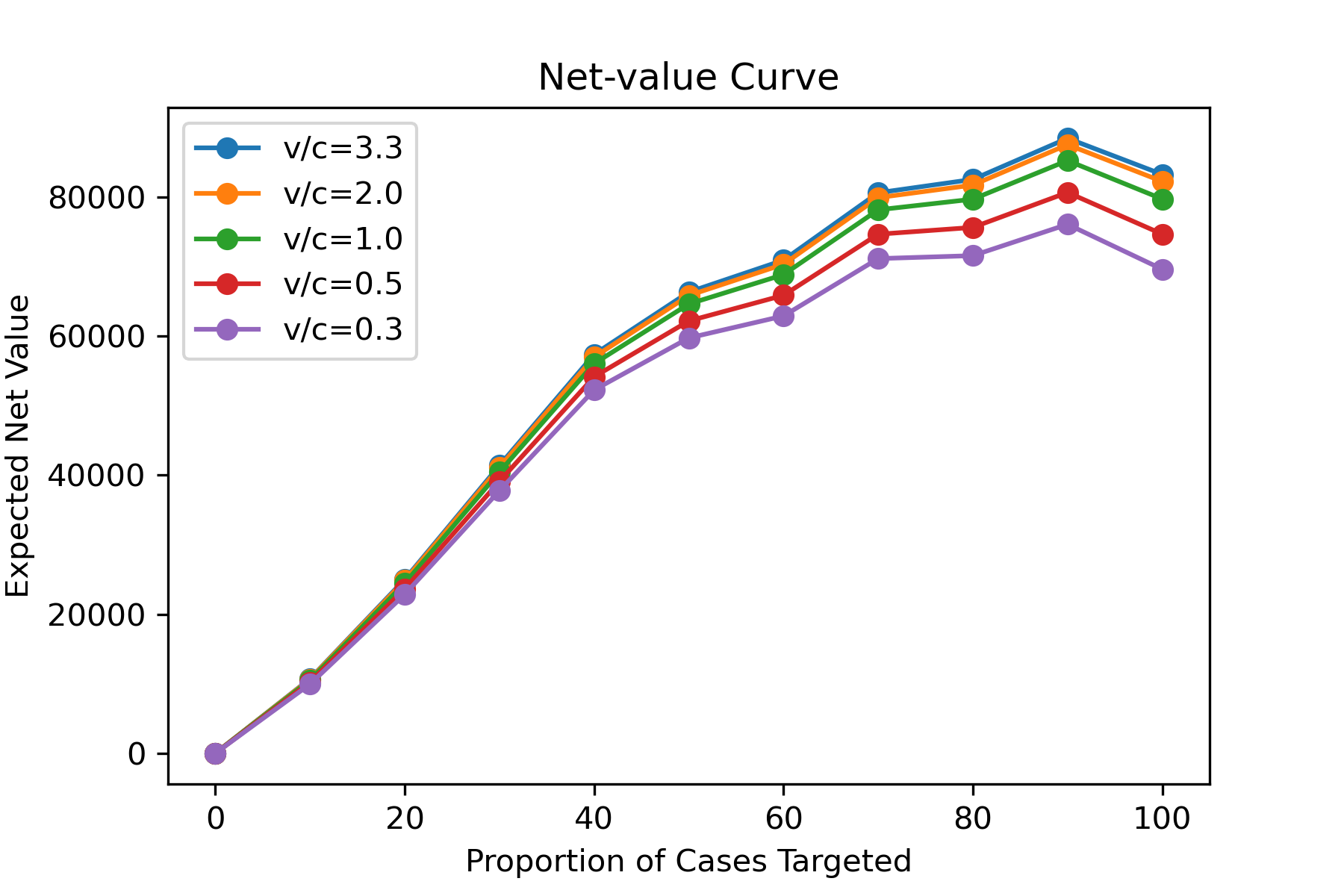}}
	\caption{Net value curves for ORF models.}
	\label{fig:NetValue}
\end{figure}

The Qini curves are without cost and benefit multipliers. So, we proceeded to plot the net value curves for both models with varying values of $v/c$, where $v$ is the value of reducing the cycle time by one day, and $c$ is the cost of applying the treatment to one case. Rather than the absolute values of $v$ and $c$, it is the ratio $v/c$ that affects the shape of the curve. Fig. \ref{fig:NetValue} shows the net value curves with different values of $v/c$ for the two logs. We observe that as the ratio $v/c$ decreases --- i.e.\ the treatment becomes more expensive relative to the benefit it provides --- the net value of treating the cases decreases, and so, it becomes more important to apply the treatment with a more targeted approach. Particularly, in the BPI17 log, the value of applying the treatment decreases more quickly than in the BPI19 log. For example, when $v/c=0.3$ the net value of applying the treatment to all cases is close to zero and the highest benefit is gained if half of the cases are treated. But in the BPI19 log the highest gain comes with treating 90\% of cases regardless of the value of $v/c$, and treating all cases still provides a gain even when $v/c=0.3$. This is because the cases in the former log are shorter and the treatment has less effect, on average, than in the latter log. Indeed, the mean duration in the BPI17 is $20$ days and the average of the estimated effects is $1$ day, while in the BPI19 log the mean duration is $150$ days and the mean treatment effect is $35$ days.





\subsection{Sensitivity Analysis}
\label{subsec: sensitivity}
\noindent Like other treatment effect estimation methods, the ORF algorithm works under the assumption of no unobserved confounding. However, when unobserved confounders do exist, our treatment effect estimates might have some bias. So, one threat to the internal validity of our results is the possibility of the presence of unobserved confounders. However, the effects of these unobserved confounders might be mild enough not to change our inference. We perform sensitivity analysis to determine how strong the effects of a confounder need to be in order for our model to change fundamentally. To achieve this, we used the Austen plots sensitivity analysis method by Veitch and Zaveri \cite{DBLP:conf/nips/VeitchZ20}. We chose this approach because it fully separates sensitivity analysis and modelling of the observed data, which allows us to use it with any machine learning method. Furthermore, this approach provides an interpretable model of the influence of unobserved confounders, which quantifies the sensitivity of our models to unobserved confounding. 




\begin{figure}[!h]
	\vspace{-\baselineskip}
	\centering
	\subfloat[BPI 2017]{\label{fig:FA1}\includegraphics[width=0.45\textwidth,height=4.4cm]{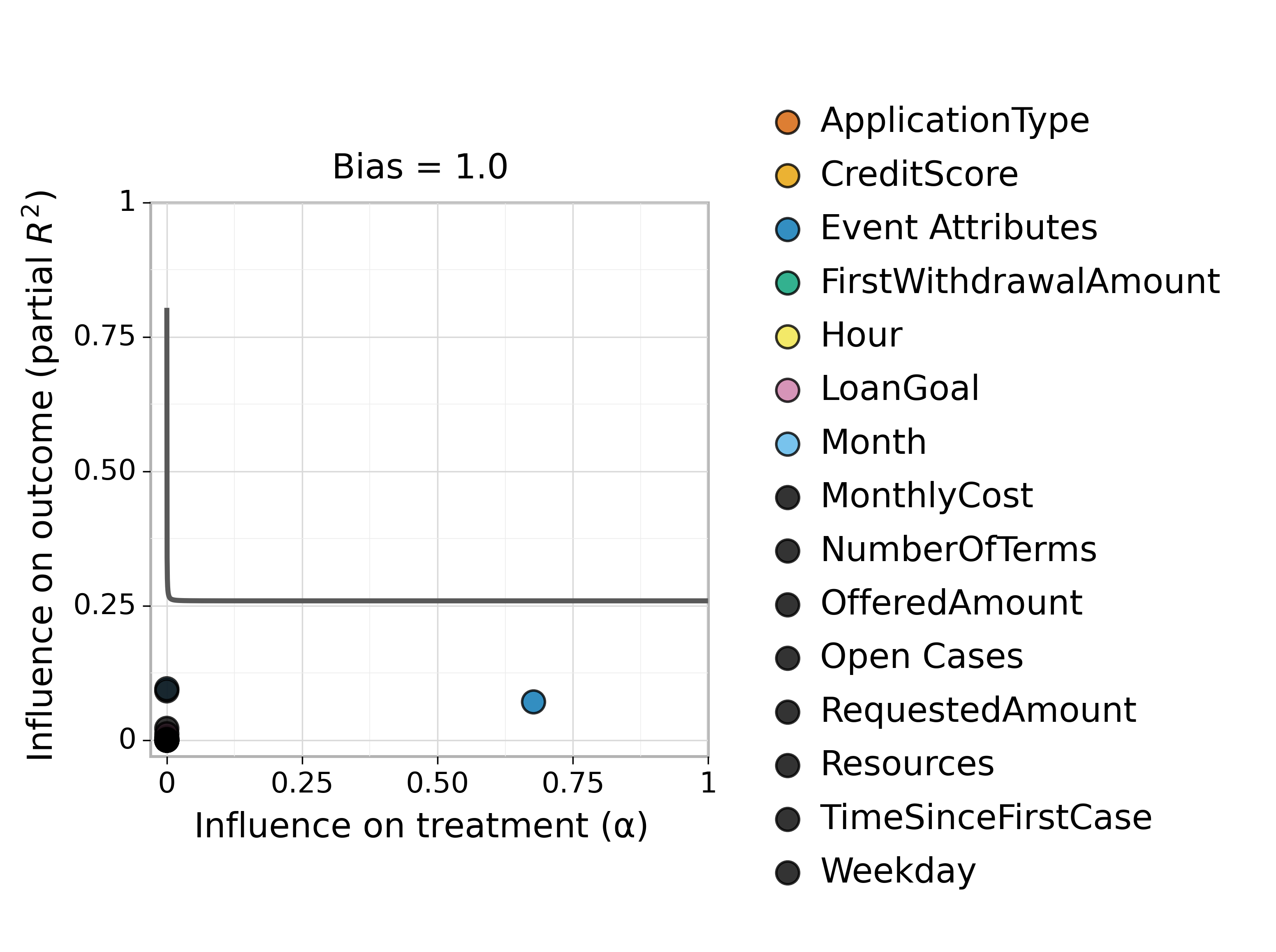}}
	\\
	\vspace{-4mm}
	\subfloat[BPI2019]{\label{fig:FA2}\includegraphics[width=0.45\textwidth,height=4.4cm]{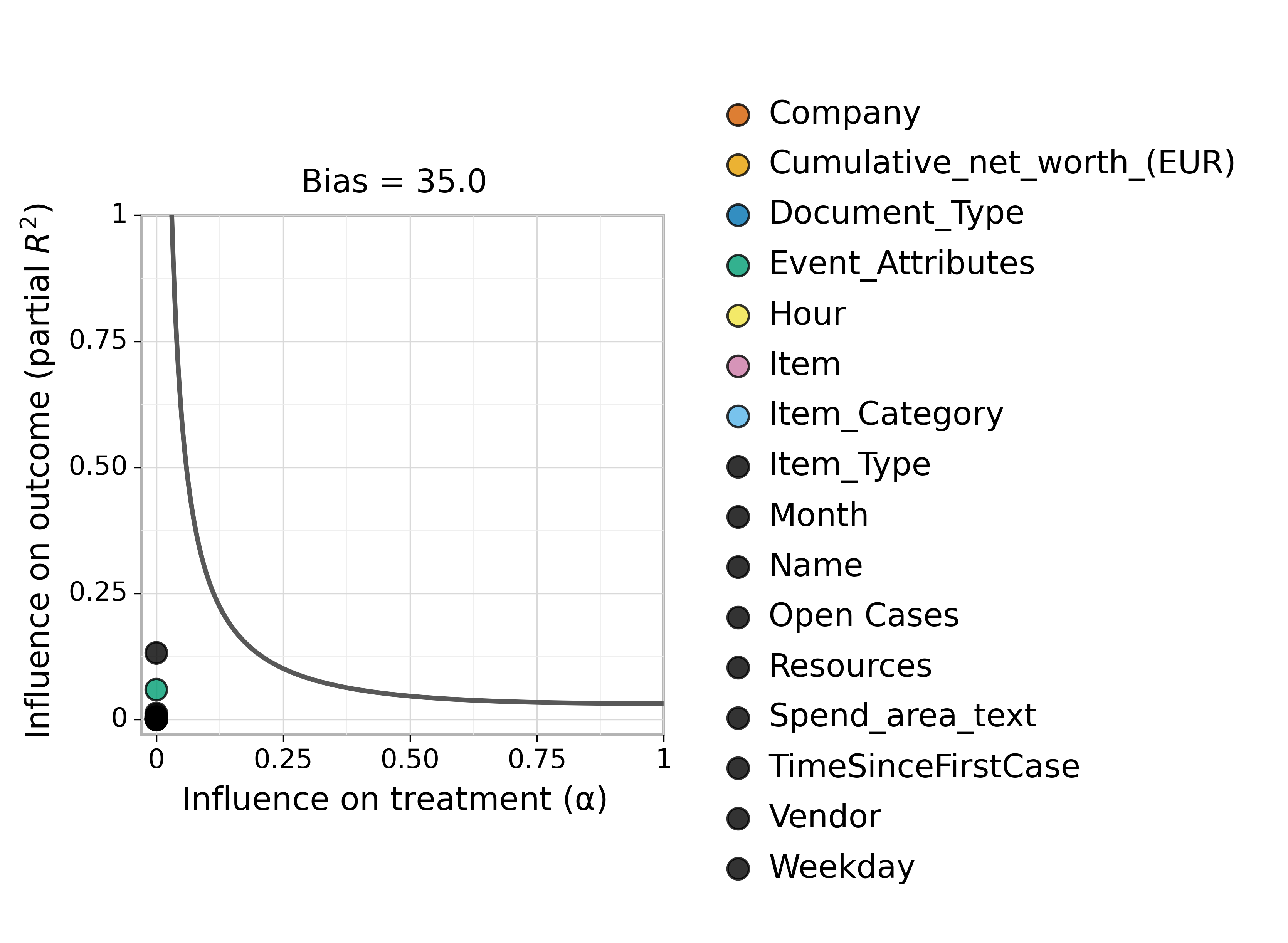}}
	\caption{Sensitivity curves for ORF models.}
	\label{fig:AustenPlot}
\end{figure}
\vspace{-2mm}

\noindent Figure \ref{fig:AustenPlot} shows the Austen plots for our models for BPI17 and BPI19 logs. The solid black line is the sensitivity curve that indicates values of $\alpha$, influence on treatment and partial $R^2$ influence on the outcome. This means that if an attribute has $\alpha$ and partial $R^2$ values above this curve, it would be sufficient to induce the bias value shown in the plot. The coloured dots show the influence of the observed potential confounders. The plots show that our conclusions from these models are robust to unobserved confounding. Since the coloured dots are well below the sensitivity curve, even the observed confounders do not have sufficient power to change the qualitative conclusions of our experiments. Particularly, the curve for the BPI17 log is very robust to confounding since it allows for the presence of variables that have a high effect on treatment assignment if their influence on the outcome is less than $0.25$. Even if an unobserved confounder exists that fully determines the treatment policy, it would need to have an influence above $0.25$ on the outcome to induce a bias of one day in the result, which in practice is highly unlikely. The model for BPI19 is more sensitive to the presence of potential confounders if they are highly influential on either the treatment or the outcome. However, since all the observed attributes are shown to be well below the sensitivity curve, the model is still robust if the effects of the hidden confounders are similar to the measured ones. It should be stressed, however, that the presence of strong hidden confounders is always a possibility, and sensitivity analysis is constrained to the available data.


\subsection{Threats to validity} 
\noindent
The evaluation comes with a threat to external validity (lack of generalisability) as it focuses on two concrete application scenarios. This can be addressed by conducting further experiments with logs of different characteristics and from different domains. A threat to construct validity is that the experiments are based on observational data. Hence, the estimated treatment effects may not match the true causal effects. To mitigate this threat, we performed sensitivity analysis to assess the robustness of our model to unobserved confounding. A rigorous A/B test should be conducted before deploying the recommendations of our method in an operational setting. 

\section{Conclusion and Future Work}
\label{sec: conclusion}
\noindent We proposed a prescriptive monitoring method that recommends if and when to apply an intervention (treatment) to an ongoing case to decrease its cycle time. The method relies on orthogonal random forests trained on historical traces to estimate the decrease in cycle time (the treatment effect) given the current state of a case. Based on this estimate, the method calculates the expected gain of the treatment given a cost function and generates recommendations based on a user-defined policy. 
We evaluated the applicability of our approach based on two real-life logs. The evaluation showed that the proposed approach yields a higher net gain than treatment policies based on non-causal predictive approaches, as previously proposed in the literature. We also showed via sensitivity analysis that the models built on these datasets are robust against potential unobserved confounding effects.


The approach assumes a binary treatment setting, i.e.\ either a treatment of a given type is applied or not. An avenue for future work is extending this method to accommodate multiple treatments (e.g. choosing between one type of treatment or another) as well as  treatments with continuous values. Our method learns the policy for which cases to treat (whom to treat). Another direction for future work is optimising the time of treatment (when to treat). Also, the approach requires the treatment to be pre-defined. Another direction for future work is to design methods to automatically discover candidate treatments from a historical log.

\smallskip\noindent\textbf{Reproducibility} The source code of our tool can be found at \url{https://github.com/zahradbozorgi/prescriptiveMonitoringORF}.

\smallskip\noindent\textbf{Acknowledgments} Research funded by the Australian Research Council (grant DP180102839) and the European Research Council (PIX Project). Thanks to Simon Remy, Kiarash Diba, and Luise Pufahl for providing preprocessed datasets.



\bibliographystyle{IEEEtran}
%
\bibliography{mybibfile2}

\end{document}